\documentclass[journal]{IEEEtran}
\usepackage{amsmath,amsfonts}
\usepackage{algorithm}
\usepackage{array}
\usepackage[caption=false,font=normalsize,labelfont=sf,textfont=sf]{subfig}
\usepackage{textcomp}
\usepackage{stfloats}
\usepackage{url}
\usepackage{verbatim}
\usepackage{graphicx}
\usepackage{cite}
\usepackage{hyperref}
\usepackage{algpseudocode}
\usepackage{setspace}
\usepackage{tabularx}
\usepackage{makecell}
\usepackage{commath}
\usepackage{multirow}

\begin{document}

\title{StarBASE-GP: Biologically-Guided Automated Machine Learning \\ for Genotype-to-Phenotype Association Analysis}

% \author{Author 1 , Author 2, Author 3
\author{Jose Guadalupe Hernandez\textsuperscript{\dag}, Attri Ghosh\textsuperscript{\dag}, Philip J. Freda\textsuperscript{\dag}, Yufei Meng, Nicholas Matsumoto, Jason H. Moore
\thanks{\textsuperscript{\dag}These authors contributed equally to this work.}
\thanks{Author Affiliations: Department of Computational Biomedicine, Cedars-Sinai Medical Center, West Hollywood, CA, 90069, USA.}

        % <-this % stops a space
% \thanks{This article has supplementary downloadable material available at https://doi.org/XX.XXXX/TEVC.XXXX.XXXXXXX, provided by the authors.}
% \thanks{Digital Object Identifier XX.XXXX/TEVC.XXXX.XXXXXXX}
}


\IEEEpubid{0000--0000/00\$00.00~\copyright~2025 IEEE}

\maketitle
% A list of 2-5 keywords are required for all manuscripts. Do not use fonts smaller than 10pt. Provide an abstract of 100-200 words that is an informative summary of the paper, including any important results found or conclusions drawn. Abstracts must not contain abbreviations, acronyms, footnotes, mathematical formulas or references. Excessively long abstracts or abstracts that contain improper material will result in paper rejection.
\begin{abstract}
We present the Star-Based Automated Single-locus and Epistasis analysis tool – Genetic Programming (StarBASE-GP), an automated framework for discovering meaningful genetic variants associated with phenotypic variation in large-scale genomic datasets. 
StarBASE-GP uses a genetic programming–based multi-objective optimization strategy to evolve machine learning pipelines that simultaneously maximize explanatory power ($r^2$) and minimize pipeline complexity. 
Biological domain knowledge is integrated at multiple stages, including the use of nine inheritance encoding strategies to model deviations from additivity, a custom linkage disequilibrium pruning node that minimizes redundancy among features, and a dynamic variant recommendation system that prioritizes informative candidates for pipeline inclusion. 
We evaluate StarBASE-GP on a cohort of \textit{Rattus norvegicus} (brown rat) to identify variants associated with body mass index, benchmarking its performance against a random baseline and a biologically naïve version of the tool. 
StarBASE-GP consistently evolves Pareto fronts with superior performance, yielding higher accuracy in identifying both ground truth and novel quantitative trait loci, highlighting relevant targets for future validation. 
By incorporating evolutionary search and relevant biological theory into a flexible automated machine learning framework, StarBASE-GP demonstrates robust potential for advancing variant discovery in complex traits.
\end{abstract}


\section{Introduction}
\label{sec:intro}
\IEEEPARstart{D}{uring} the \textit{Modern Synthesis}, Ronald Fisher and Sewall Wright developed a mathematical framework for decomposing variation in observable traits (phenotypes) in living systems\cite{fisher1919xv,wright1931evolution}.
Their framework quantified the relative contributions of genetic (\( V_G \)) and environmental (\( V_E \)) factors to overall phenotypic variance (\( V_P \)):
\[ V_P = V_G + V_E \]
Later, Douglas Falconer and Trudy Mackay incorporated genotype-by-environment interactions (\( V_{G \times E} \)) into this framework, extending phenotypic variance to account for differential genotypic responses to changing environments\cite{falconer1983quantitative}:
\[V_P = V_G + V_E + V_{G \times E}\]
The genetic component (\( V_G \)) can be further decomposed into three sources: additive variance (\( V_A \)), which represents the cumulative effects of individual alleles; dominance variance (\( V_D \)), which arises from non-additive interactions between alleles at the same locus; and epistatic variance (\( V_I \)), which captures interactions between alleles at different loci \cite{fisher1919xv,wright1931evolution}.
With these three additional components, the full decomposition of phenotypic variance becomes: 
\[V_P = (V_A + V_D + V_I) + V_E + V_{G \times E} \]

Understanding and predicting the relationship between genotype and phenotype is a fundamental goal in genetics, aiming to resolve the genetic components ($V_A + V_D + V_I$) underlying phenotypic variance ($V_P$). 
Genotype-to-phenotype association (GPA) methods aim to quantify how genetic variation ($V_G$) contributes to observable traits and diseases\cite{falconer1983quantitative, lynch1998genetics}.
Genome-wide association studies (GWAS) and quantitative trait locus (QTL) analysis have become foundational tools in GPA, enabling the identification of thousands of loci linked to traits and diseases\cite{uffelmann2021genome, myles2008quantitative}. 

\IEEEpubidadjcol

GPA methods typically rely on linear models, where biallelic genotypes are encoded as $0$, $1$, and $2$ to represent the number of minor alleles. 
This additive encoding assumes heterozygote ($1$) phenotypes lie exactly between the two homozygotes classes ($0$ and $2$)\cite{falconer1983quantitative, lynch1998genetics}. 
While convenient and interpretable, the additive encoding limits the detection of dominance effects ($V_D$), where genotypes deviate from additive patterns. 
In addition, standard GWAS and QTL frameworks rarely account for epistasis ($V_I$).
These non-linear interactions obscure causal relationships and dramatically expand the search space, rendering them difficult to detect with traditional association models\cite{moore2009epistasis, batista2024interaction}. 
As a result, most GWAS and QTL studies systematically overlook non-additive variation, leading to incomplete effect estimates that contribute to the broader issue of \textit{missing heritability}\cite{tam2019benefits}.

Mounting evidence suggests that non-additive effects significantly contribute to trait variation in both model systems and humans\cite{wermter2008preferential, hallin2016powerful, wu2016comparative, yang2017incomplete, bonnafous2018comparison, liu2021recessive, matsui2022interplay, freda2024pager}. 
Notably, Matsui \textit{et al.} (2022) found that dominance and epistatic effects accounted for one-third of broad-sense heritability across multiple traits and environments in diploid yeast (\textit{Saccharomyces cerevisiae}), underscoring the importance of modeling non-additive variation to resolve the genetic architectures of complex traits.

GWAS and QTL analyses are also limited by their exclusive reliance on \textit{p}-values to infer associations~\cite{myles2008quantitative, uffelmann2021genome}.
Variants are ranked, and significance thresholds are determined either empirically (\textit{e.g.}, permutation testing) or conventionally, as with the widely used ``\textit{\(\mathit{5 \times 10^{-8}}\) cutoff}" in human GWAS studies\cite{fadista2016famous}.
These thresholds are often arbitrary and may not reflect a clear biological distinction between causal and non-causal variants.
Additionally, \textit{p}-values are influenced by numerous factors, including sample size, population structure, cohort composition, allele frequencies, multiple testing corrections, and linkage disequilibrium (LD)\cite{johnson2010accounting, hong2012sample, fadista2016famous}.
Nevertheless, most GPA methods rely on these criteria to select putative targets, despite their limitations.

Multiple testing correction presents a major challenge for GPA methods, which face the \textit{``small \(n\), large \(p\)"} problem where the number of genetic variants far exceeds sample size\cite{diao2013assessing}. 
To control type I error inflation, stringent significance thresholds are applied, often increasing type II error rates\cite{lambert2012learning, uffelmann2021genome}. 
Consequently, variants that fail to reach genome-wide significance may still be functionally relevant. 
This prevents researchers from identifying a definitive set of variants for downstream analysis, forcing reliance on prioritization beyond statistical thresholds.
Several approaches address this issue by reducing the number of variables ($p$): stepwise regression iteratively selects predictive variants\cite{cordell2002unified}, while LASSO and ridge regression apply regularization penalties\cite{malo2008accommodating, wang2011identifying}. 
However, these methods are computationally intensive, especially when extended to model additional components of $V_G$, such as alternative inheritance models or epistatic interactions.

Collectively, these challenges hinder the prioritization of variants for replication and downstream functional validation, limiting translational impact. 
Our goal is to overcome these limitations by leveraging machine learning (ML) and genetic programming (GP) to build a GPA framework that models both $V_A$ and $V_D$, adaptively explores high-dimensional search spaces, and evaluates variant importance beyond traditional statistical limitations.
Here, we introduce StarBASE-GP (Star-Based Automated Single-locus and Epistasis analysis tool – Genetic Programming), an automated machine learning (AutoML) package specifically designed to enhance GPA analysis through the integration of biological domain knowledge.
\section{Related Works}
Machine Learning (ML) methods have emerged as powerful tools for GPA, particularly for detecting non-additive effects\cite{nicholls2020reaching, sigala2023machine, yao2013random}.
Random Forest approaches improve variant selection by reducing dimensionality and enhancing interaction detection in datasets with high dimensionality\cite{nguyen2015genome}. 
Gradient boosting methods have been used to capture non-linear variant interactions, facilitating the identification of risk-predictive variant groups in breast cancer while outperforming traditional models\cite{behravan2018machine}.
Beyond variant discovery, ML has also played a key role in post-GPA analysis, identifying putative targets from highly correlated variants in LD\cite{nicholls2020reaching}.
These methods enhance functional interpretation while addressing some of the limitations of traditional statistical approaches.

Many ML approaches, either internally or externally, generate feature importance (FI) metrics that quantify the predictive contribution of each feature within a model \cite{breiman2001random, lundberg2017unified}. 
While FI metrics are not directly comparable to \textit{p}-values, they offer intuitive and interpretable model-specific insights into predictive power. 
Unlike \textit{p}-values, which typically assess the probability of an effect under a null hypothesis for each variable in isolation, FI metrics inherently account for other features\cite{molnar2020interpretable}.
This distinction is relevant in complex, high-dimensional data, where interactions and dependencies between features are prevalent \cite{moore2009epistasis}.
FI methods help capture these non-linear relationships without specific data assumptions\cite{lundberg2017unified}, and provide a continuous measure of predictive impact without relying on arbitrary significance cutoffs\cite{breiman2001random}. 
Rather than estimating probabilities of true associations, methods such as Shapley values\cite{lundberg2017unified}, permutation importance\cite{altmann2010permutation, breiman2001random}, and Gini importance\cite{nguyen2015genome} evaluate the relative influence of each feature on model predictions directly.
While FI metrics do not eliminate the need for thresholds, they provide direct, adaptable, and context-aware measures of feature influence by emphasizing predictive power over statistical inference.

While ML methods have the potential to overcome the limitations of traditional GPA approaches\cite{freda2023autoqtl, manduchi2022promise, le2020scaling, sohn2017toward, hao2022knowledge}, they introduce new challenges such as increased computational cost\cite{talaei2023machine}, susceptibility to overfitting\cite{demvsar2021hands}, and difficulties in model selection and interpretation\cite{hassija2024interpreting}.
Moreover, applying ML to genetic analysis requires domain-specific expertise, particularly in designing an ML pipeline.
For example, a geneticist investigating the genetic basis of a disease must first determine an appropriate strategy for preprocessing and encoding genetic data.
Next, they would apply feature selection strategies (\textit{e.g.}, filtering by allele frequency, statistical association, or domain knowledge) to reduce dimensionality while preserving variation.
This is followed by determining the best modeling approach--linear models for additive effects or non-linear models, like random forests\cite{nguyen2015genome}, to capture complex interactions--while ensuring proper validation to prevent overfitting and bias.
Finally, they must structure the overall pipeline by determining the number of operators at each stage and tuning hyperparameters to ensure seamless integration across preprocessing, feature selection, and modeling.
This multi-step process can be challenging for researchers lacking specialized ML expertise, especially given the vast range of pipeline configurations that can be tested.

Automated machine learning (AutoML) addresses these challenges by automating key pipeline building steps such as algorithm selection, hyperparameter tuning, bias control, and model interpretation\cite{hutter2019automated, manduchi2022promise}. 
This reduces manual trial-and-error, making ML-based GPA more accessible and efficient. 
AutoQTL, an AutoML tool introduced by Freda \textit{et al.}, 2023\cite{freda2023autoqtl}, uses genetic programming (GP) to automate decisions typically made by geneticists designing GPA studies, while incorporating principles of quantitative genetics. 
These include genotype encoding to capture dominance deviations ($V_D$), allele frequency-based variant filtering, and regression method selection for modeling single- and multi-locus effects. 

Despite its advantages, AutoQTL has notable limitations. 
It was designed primarily as a post-GPA validation tool for small variant subsets to detect non-additive effects.
Additionally, it applies the same feature encoding to all variants, limiting the ability to model variant-specific inheritance patterns observed in biological systems\cite{wermter2008preferential, hallin2016powerful, wu2016comparative, yang2017incomplete, bonnafous2018comparison, liu2021recessive, matsui2022interplay, freda2024pager}.
Furthermore, AutoQTL’s optimization relied on two correlated objectives: validation performance (pipeline $r^2$) and a custom overfitting metric defined as the performance difference between training and validation sets. 
Because validation performance appears in both, the objectives were not truly independent, limiting the algorithm’s ability to explore meaningful trade-offs.
In this work, we advance prior efforts by deepening domain knowledge integration, enabling variant-specific inheritance modeling, and supporting large-scale genomic datasets—demonstrating that AutoML can be effectively applied to GPA analysis.
\section{StarBASE-GP}
\label{sec:StarBASE-GP}
StarBASE-GP combines automated machine learning (AutoML), genetic programming (GP), Pareto optimization, and biological domain knowledge to identify meaningful genetic variants associated with complex traits in large-scale genomic datasets.
Specifically, StarBASE-GP uses a GP-based Pareto optimization framework to evolve a population of pipelines, simultaneously maximizing pipeline $r^2$ and minimizing complexity.
The design of StarBASE-GP addresses the challenges of the \textit{``small \(n\), large \(p\)"} problem by using a subset-based search strategy to handle high dimensionality efficiently, where pipelines process variant subsets rather than the entire dataset.
The GP system iteratively selects the most informative variants, assembling optimal feature sets that maximize phenotypic explainability. 
A variant's explainability is assessed by quantifying its contribution to genetic variation in terms of both additivity (\( V_A \)) and dominance (\( V_D \)), encoding each variant in its optimal inheritance model.
This information is stored and used to construct a recommendation system that guides pipeline development by filtering out suboptimal variants, maximizing phenotypic explainability and facilitating dataset exploration.
To maximize explainability (pipeline $r^2$) and dataset exploration, StarBASE-GP uses multiple biologically-inspired mechanisms:

\begin{itemize}
    \item Genomic proximity-based variant grouping: Promotes discovery by prioritizing variants from distant genomic regions, preventing the search from converging to local optima caused by linkage disequilibrium (LD) and ensuring broad dataset coverage.
    
    \item Custom LD-pruning node: Eliminates redundant variants through competitive selection, ensuring that only the most informative variants persist through pipeline evolution.
    
    \item Biologically-inspired ML feature selectors: Utilizes principles such as allele and genotype frequency to filter out variants with minimal impact on phenotypic variation.
\end{itemize}

Incorporating biological domain knowledge in ML has been shown to enhance feature selection and feature engineering, improving the predictive performance of models applied to complex problems in biology, genetics, and medicine\cite{manduchi2022promise, le2020scaling, sohn2017toward, hao2022knowledge, freda2023autoqtl}.
By integrating domain-specific heuristics, StarBASE-GP provides a scalable, biologically relevant, and computationally efficient framework for GPA.

\subsection{Maximizing data usage and information}
\label{sec:snp:data}
Efficiently identifying relevant variants is key to optimizing GPA prediction amid potentially millions of candidate variants. 
To achieve this, we implement three key strategies for extracting maximal information from given data. 
First, we assume that the genotypes and variant labels adhere to a predefined format (Section \ref{sec:data-format}). 
Second, we determine the optimal encoding for each variant to maximize its explainability and contribution to predictive performance (Section \ref{sec:snp_encoding}). 
Finally, we maintain a variant database (\textit{SNP DB}), that systematically tracks relevant information, which is later leveraged throughout the evolutionary process (Section \ref{sed:snp_db}).

\subsubsection{Data format}
\label{sec:data-format}
Single-nucleotide polymorphism (SNP) data, encoded in the standard additive format ($0$, $1$, and $2$), with a quantitative phenotype, are provided as input.
SNPs are the most common type of genetic variant used in GPA studies and are single base pair (bp) positions in which the genotype varies across individuals in a population (\textit{i.e.}, segregating alleles)\cite{falconer1983quantitative}.
The SNP dataset is divided into training and validation sets ($50$/$50$ by default).
Any missing genotype values are mode-imputed per SNP locus to ensure data completeness.
For ease of comparison to other encoding strategies and to improve interpretability, StarBASE-GP normalizes the default additive encoding between 0 and 1.

StarBASE-GP requires that each SNP column label in the dataset follows the format $chr.position$, where $chr$ specifies the numeric chromosome on which the SNP is located, and $position$ indicates the numerical bp coordinate (location) within that chromosome.
This format allows a comprehensive scan of the entire genome to extract positional information leveraged during the evolutionary process.
For example, this allows SNPs to be grouped by chromosome or genomic proximity to remove redundancies introduced by LD.
LD typically occurs when genetic variants in close proximity on a chromosome are highly correlated due to their tendency to be inherited together during recombination (meiosis)\cite{falconer1983quantitative}.

Each SNP in the dataset is assigned to a \textit{bin} by grouping neighboring SNPs on chromosomes. 
SNPs are first sorted by bp position.
Binning then begins with the SNP at the smallest bp position (most upstream), and SNPs are sequentially added until the user-defined \textit{Bin Size} is reached.
A new bin is then created and the process continues until all SNPs are assigned. 
This binning strategy enables targeted exploitation and broad exploration of the genome by prioritizing SNPs that are ``in bin", ``out of bin", or ``out of chromosome".
Collectively, these parameters form a search strategy spectrum: ``in bin" identifies ``\textit{peaks}" of phenotypic explainability (exploitation), while ``out of chromosome" encourages broader discovery across the genome (exploration).

\subsubsection{SNP encoding}
\label{sec:snp_encoding}
In standard GPA, SNPs are encoded under a single inheritance model.
This approach assumes that a single model can adequately capture additive and dominance effects across all SNPs.
However, the phenotypic distributions per genotypic class ($0$, $1$, and $2$) can follow unique inheritance models for each SNP\cite{falconer1983quantitative, lynch1998genetics}.
Therefore, this is no ``\textit{one size fits all}'' strategy to inheritance model encoding.

\begin{figure}[!h]
\centering
\includegraphics[width=\linewidth]{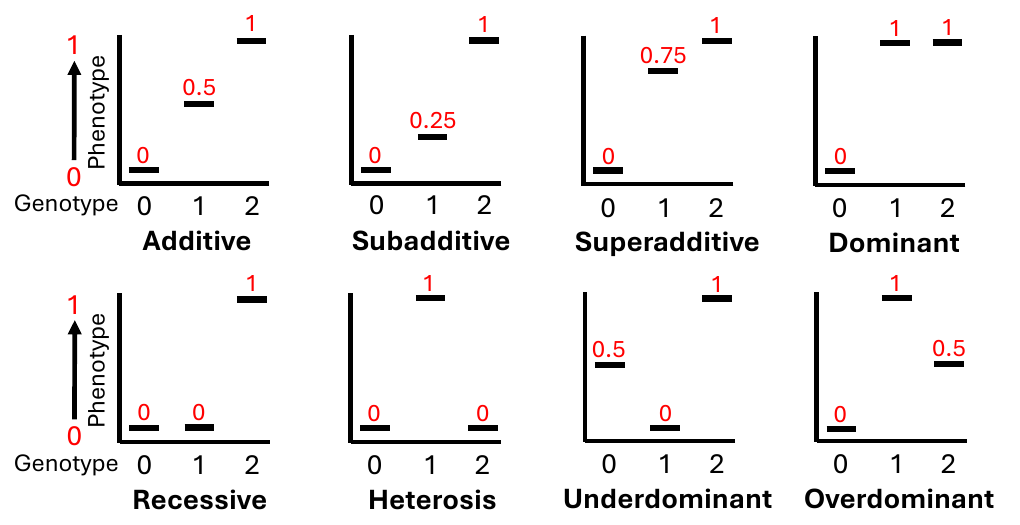}
\caption{The eight ``\textit{strict}" inheritance models used in StarBASE-GP. On the \textit{x}-axes are the three typical genotype encodings found in SNP datasets ($0$, $1$, and $2$: additive). On the \textit{y}-axes is the expected phenotype increasing from $0$ to $1$. The red numbers above horizontal bars within plots represent the StarBASE-GP genotype encodings enforced under each respective inheritance model.}
\label{fig:inheritance}
\end{figure}

In StarBASE-GP, each SNP is assigned its optimal inheritance encoding when first incorporated into a pipeline, whether during population initialization or subsequent evolution.
This encoding is selected by evaluating eight different inheritance models (additive, superadditive, subadditive, dominant, recessive, heterosis, overdominant and underdominant (Figure \ref{fig:inheritance})), which capture a broad range of inheritance patterns observed in living systems\cite{wermter2008preferential, hallin2016powerful, wu2016comparative, yang2017incomplete, bonnafous2018comparison, liu2021recessive, matsui2022interplay, freda2024pager}. 
Since GPA models rely on relative phenotypic differences between genotypic classes ($0$, $1$, and $2$), rather than absolute values \cite{draper1981applied}, an inheritance model and its mirrored equivalent are functionally identical in terms of model fit.
For instance, if a SNP follows the phenotypic pattern $1$, $0$, $0$ instead of $0$, $1$, $1$, dominant encoding ($0, 1, 1$) will still be optimal, as genotypes $1$ and $2$ remain equivalent relative to genotype $0$. 
This scale invariance ensures that encoding selection is consistent, independent of phenotype direction (Supplemental Information).

A ninth encoding strategy, Phenotype Adjusted Genotype Encoding and Ranking (PAGER), is also included. 
PAGER, unlike the other eight encodings, does not assume a strict inheritance model.
Instead, PAGER calculates the normalized (between $0$ and $1$) mean phenotypic differences between genotypic classes, which replace the original additive encodings\cite{freda2024pager}. 
In this way, PAGER can encode inheritance patterns not captured by the other ``\textit{strict}" models.

For each SNP in a pipeline, the encoding that yields the greatest validation marginal $r^2$ from a linear regression of the phenotype on the genotypes in the training set (\textit{train} in Algorithm \ref{alg:nsga}) is selected as the optimal encoding.
This encoding replaces the additive encoding in the validation set (\textit{validate} in Algorithm \ref{alg:nsga}), unless the additive model achieves the greatest marginal validation $r^2$.
The greatest marginal validation $r^2$ is then used to inform feature selection and recommendation during the evolutionary process. 
We use marginal $r^2$ because it quantifies the proportion of phenotypic variance explained ($V_P$) by a single SNP.
Additionally, as it is unit-less, it allows for an objective selection of the encoding that best captures the SNP’s contribution to the phenotype\cite{chicco2021coefficient}.

When a SNP is first incorporated into a pipeline, the marginal validation $r^2$ and its corresponding encoding are stored for later use in SNP DB to eliminate redundant calculations, as these values are inherent properties of a SNP that are not affected by other SNPs in a pipeline.
If a SNP has a negative marginal validation $r^2$ under its optimal encoding, it is deemed uninformative\cite{chicco2021coefficient} and is excluded from future pipelines.
This strategy allows pipelines to contain multiple SNPs under various inheritance encodings that maximize their respective explainabilities while collectively improving the performance of the entire pipeline.

\subsubsection{SNP database (SNP DB)}
\label{sed:snp_db}
We record key information for each SNP to identify useful SNPs for further exploration:

\begin{itemize}
    \item Marginal validation $r^2$: The regression $r^2$ value for a SNP obtained under its optimal inheritance encoding. This value captures a SNP’s relevance for identifying new candidates in future pipelines. It is also used as a comparison metric for potentially correlated SNPs in the LD-pruning node. By default, this value is \textit{None}.
    
    \item Encoding strategy: The optimal encoding selected from the nine possible inheritance models. This value transforms the SNP into its most effective representation for analysis in a pipeline. By default, this value is \textit{None}.
    
    \item Bin ID: The identifier of the chromosome and bin to which a SNP belongs. This value represents a SNP's relative position within the genome and is used for SNP recommendations during pipeline construction.
    
    \item Consideration flag: A Boolean flag indicating whether a SNP remains recommendable. The flag is set to \textit{false} if the SNP exhibits a negative marginal $r^2$ (under all encodings) or has not survived its pipeline LD-pruning node. By default, the flag is set to \textit{true}.
\end{itemize}

\subsection{Biologically inspired pipeline structures}
StarBASE-GP pipelines incorporate biological nuances typically considered by biologists, enabling the efficient and effective processing of high-dimensional datasets.
Each pipeline is allocated a subset of SNPs, with the number ranging from one to a user-defined maximum.
Once encoded, SNPs undergo an initial round of feature selection via LD pruning.
The SNPs that remain after LD pruning undergo an additional round of feature selection via conventional ML or biologically inspired methods.
This filtering order enables the LD pruning component to eliminate redundant SNPs with traditional biological practices, allowing the subsequent feature selection process to focus on identifying the most informative SNPs.
The SNPs that survive feature selection are then fed into an ML regressor for training and prediction.

\begin{figure}[!h]
\centering
\includegraphics[width=\linewidth]{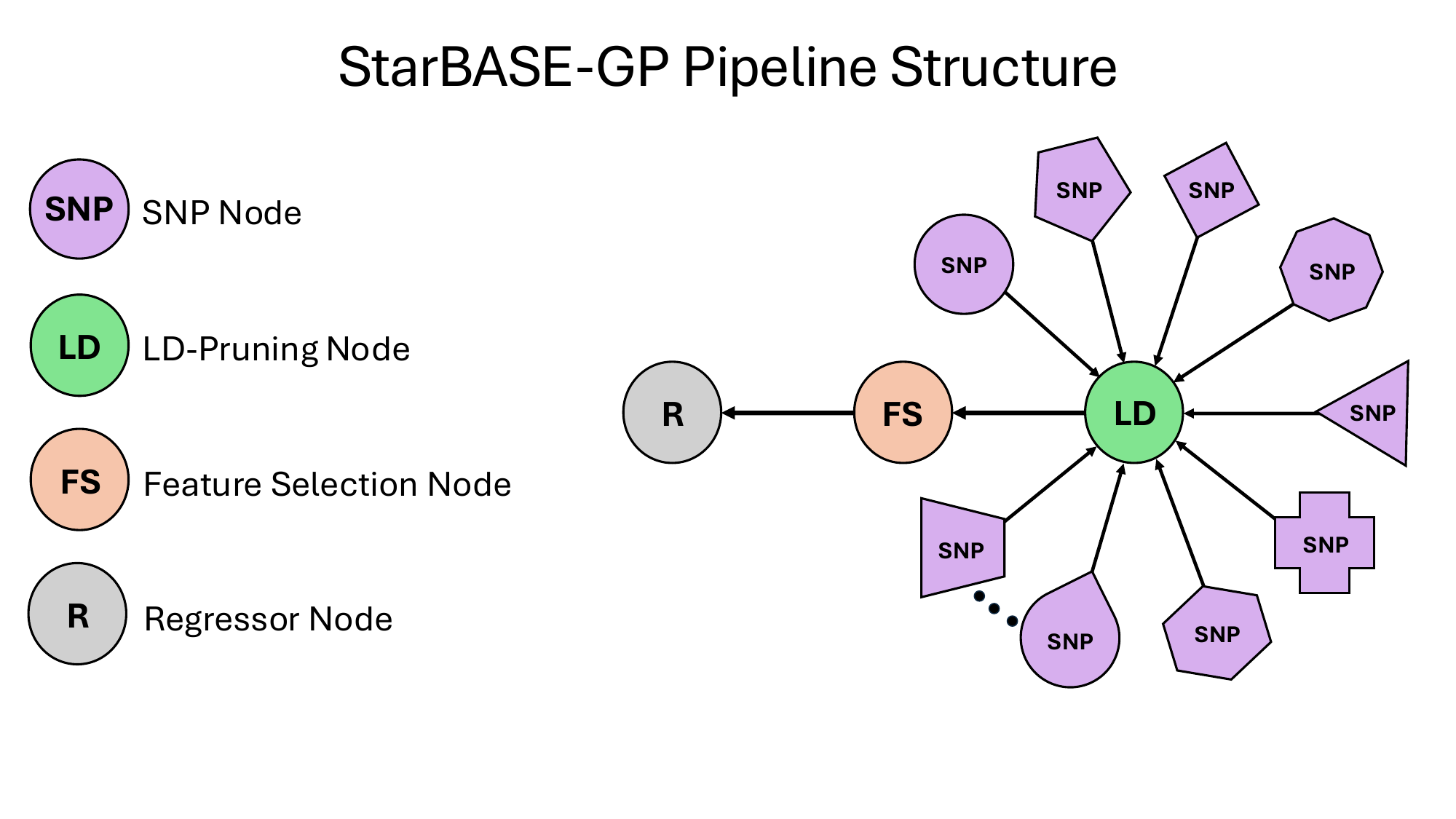}
\caption{StarBASE-GP pipeline structure. The various shapes for SNP nodes represent different inheritance model encodings. Colored circles in the legend (left) describe the node types each pipeline contains. All StarBASE-GP pipelines follow this defined structure. However, pipelines differ in their node hyperparameters, as well as in the number and composition of SNP nodes.}
\label{fig:pipeline}
\end{figure}

Figure \ref{fig:pipeline} illustrates the pipeline structure used in StarBASE-GP, represented as a directed acyclic graph.
In this representation, nodes encapsulate algorithms and their corresponding hyperparameters, while directed edges indicate the flow of information between nodes.
The nodes within a pipeline are classified into one of four categories--(1) SNP, (2) LD-pruning, (3) feature selection, or (4) regressor--and process information sequentially in this order.
Each node executes an algorithm that corresponds to its designated category.

\subsubsection{SNP node}
\label{sec:SNP_node}
Each SNP is processed by a pipeline and transformed to its optimal encoding (Section \ref{sec:snp_encoding}).
Each pipeline is ensured to process a distinct set of SNPs, preventing redundant SNP insertions. 
However, the same SNP may be included in multiple pipelines, and pipelines may share identical or overlapping subsets of SNPs.
Users specify the maximum number of SNPs that a pipeline can process.
This constraint must ensure that the selected SNP range remains compatible with the ML regressors used in this study, adhering to the general recommendation that the number of samples should far exceed (usually by $10x$) the number of features to maintain robust model performance\cite{peduzzi1996simulation}.

\subsubsection{LD-pruning node}
\label{sec:ld_prune_node}
Standard LD pruning approaches typically apply liberal correlation thresholds, eliminating only highly linked variants. 
Thus, substantial residual correlation among the remaining variants can persist, increasing search complexity and introducing redundancy into modeling efforts. 
To further disentangle independent genetic signals, additional post-processing steps are often required.
One such approach is conditional analysis, which identifies independently associated variants by assessing their effects while accounting for correlations with other variants\cite{yang2012conditional}. 
This method is performed to distinguish true genetic associations from those driven by LD, ensuring that retained variants contribute meaningful, independent signals in downstream modeling. 

We implement an adaptive process that evolves an optimal LD threshold using the coefficient of determination (\(r^2\)) as the correlation metric.
This threshold is evolved as a hyperparameter between default values of $0.50$ and $0.95$ in shifts of $0.05$, covering a continuum from moderate to liberal pruning.
To assign SNPs to LD groups, StarBASE-GP identifies all SNPs from the same chromosome in a pipeline and sorts by bp location. 
The algorithm then processes this ordered list and builds LD groups sequentially using a grouping threshold ($d_{max}$), which is evolved as a hyperparameter ranging from default values of $500,000$ bp to $1,000,000$ bp in $100,000$ bp shifts.
It begins a group with the leading upstream SNP, and appends every subsequent SNP whose distance from the previous satisfies $ \abs{bp_{i} - bp_{i-1}} \leq d_{max}$. 
When the bp distance between adjacent SNPs exceeds $d_{max}$, the group is closed and a new one is started.
Groups are then passed to LD-pruning.

For each LD group, we compute a correlation matrix containing the \(r^2\) for every pairwise SNP combination, where \(r^2\) represents the squared correlation coefficient based on the additive encoding in the training set.
The additive encoding preserves allele counts, which are necessary for LD assessment\cite{falconer1983quantitative, lynch1998genetics}. 
If the correlation of any pair exceeds the evolved \(r^2\) threshold, we compare their marginal validation \(r^2\) values (under the optimal encodings) and retain the SNP with the higher value.
Any SNP removed during pairwise comparisons is discarded from subsequent correlation checks.

All retained SNPs within their respective LD groups are then passed to conditional analysis using the Wald test.
This test assesses whether a SNP has a meaningful phenotypic effect ($\beta$ significantly different from zero) after adjusting for other SNPs in the group\cite{fahrmeir2022regression}. 
Within each LD group, the SNP with the highest marginal validation \(r^2\) (under its optimal encoding) is designated as the anchor SNP.
Each non-anchor SNP remaining in the group is then tested (also under its optimal encoding) in a regression model where the phenotype is regressed on the non-anchor SNP with the anchor SNP as a covariate using the training set.
If the Wald test statistic is significant (\textit{p} $<$ $0.05$ as a default) after multiple test correction (adjusts for the number of Wald tests performed in the group using Benjamini-Hochberg FDR), the non-anchor SNP is retained; otherwise, it is discarded.
This process is performed for each LD group, ensuring that only the most informative (independent) SNPs remain. 
If an LD group contains only one SNP before conditional analysis, that SNP is retained.

After LD-pruning and conditional analysis, all SNPs discarded during the pruning process are sent to update the SNP DB (Section \ref{sed:snp_db}).
Critically, the LD-pruning node not only streamlines feature selection but also combines traditionally separate pre- and post-processing steps into a single, integrated operation.
By enforcing a structured selection process, SNPs are evaluated within their respective genomic regions, ensuring that only the most important and informative peaks survive to continue in the evolutionary process. 
This adaptive approach not only reduces the search space but also enhances the identification of SNPs that best describe the genetic architecture of the phenotype.
The LD-pruning node allows each pipeline to transfer knowledge of informative SNPs, guiding the evolutionary process toward increasingly optimal pipelines.

\subsubsection{Feature selection node}
The SNPs that pass the LD-pruning node serve as representative subsets of specific genomic regions, capturing the most informative features assigned to the pipeline.
The feature selection node is crucial because some SNPs of low importance may pass LD-pruning due to inadequate sampling or incomplete linkage information. 
Thus, feature selection provides additional filtering to identify the most informative SNPs.

The feature selection nodes implement one of seven algorithms: six algorithms from scikit-learn\cite{pedregosa2011scikit} or a custom method inspired by biological principles.
The six scikit-learn feature selection techniques include: (1) variance threshold, (2) select percentile, (3) family-wise error control, (4) L1 regularization, (5) tree-based feature selection, and (6) tree-based sequential feature selection. 
The biologically inspired feature selection algorithm (7) selects features based on predefined thresholds for genotype and allele frequencies, as features with low variation may be less informative, lack statistical power, and contribute to spurious associations, reducing their reliability in downstream analyses \cite{lynch1998genetics, crow2017introduction}.

\subsubsection{Regressor nodes}
The SNPs that reach the regressor node have survived both phases of feature selection and their count define a pipeline's \textit{complexity}.
These SNPs are fed into the regressor for training and prediction, yielding a pipeline's $r^2$ performance.
The algorithms used in regressor nodes are sourced from the scikit-learn package \cite{pedregosa2011scikit}.
Specifically, each regressor node uses one of seven regressors, along with corresponding hyperparameters: (1) linear regression, (2) linear regression with L1 and L2 regularization, (3) linear regression with stochastic gradient descent, (4) support vector regression, (5) decision tree regression, (6) random forest regression, and (7) gradient boosting regression.

\subsection{Pareto optimization through genetic programming}
\label{sec:engine}
StarBASE-GP implements a custom non-dominated sorting genetic algorithm II (NSGA-II) \cite{deb2002nsgaii} to evolve a population of pipelines that approximate a Pareto front, simultaneously maximizing pipeline $r^2$ and minimizing complexity.
Finding the Pareto front for these conflicting objectives enables the identification of optimal SNP subsets, as any pipeline on the front cannot further improve $r^2$ or reduce complexity without worsening the other objective.
By using Pareto optimization, a pipeline that includes a single informative SNP is safeguarded against pipelines containing a larger number of SNPs that inflate pipeline $r^2$, preserving smaller pipelines and enabling the propagation of valuable variants to future pipelines.

\begin{algorithm}
\setstretch{1.05}
\caption{StarBASE-GP Algorithm}
\label{alg:nsga}
\begin{algorithmic}[1]
\Procedure{Evolve}{$size, gens, train, validate$}
    \State pop = \textit{Population\_Initialization}($size$)
    \State \textit{Process\_SNP}(pop,SNP\_DB,$train$,$validate$)
    \State \textit{Pipeline\_Evaluation}(pop,$train$,$validate$)
    \State \textit{Pruned\_SNP}(pop,SNP\_DB)
    \For{gen = 1 to $gens$}
    \State parents = \textit{Parent\_Selection}(pop) 
    \State offspring = \textit{Offspring\_Generation}(parents,$2*size$)
    \State \textit{Process\_SNP}(offspring,SNP\_DB,$train$,$validate$)
    \State \textit{Pipeline\_Evaluation}(offspring,$train$,$validate$)
    \State \textit{Pruned\_SNP}(offspring,SNP\_DB)
    \State pop  = \textit{Survival\_Selection}(offspring,$size$)
    \EndFor
\State \textbf{return} pop
\EndProcedure
\end{algorithmic}
\end{algorithm}

Before starting, the user provides the number of generations to evolve the initial population ($gens$), the number of pipelines to retain at each generation ($size$), and the genetic dataset being analyzed, which will be evenly split into a training set ($train$) and a validation set ($validate$).
Once these variables are set, the initial population is constructed (Algorithm \ref{alg:nsga}, Line 2).
The initial population comprises randomly constructed pipelines with the user-defined maximum SNPs per pipeline.
Remaining nodes are constructed by randomly assigning both node algorithms and sampled sets of hyperparameters.

After constructing the initial population, we update SNP DB (Section \ref{sed:snp_db}) with the optimal encoding strategy and corresponding marginal validation $r^2$ for each assigned SNP (Algorithm \ref{alg:nsga}, Line 3). 
Once SNP DB is updated, the pipelines in the initial population are evaluated to compute their $r^2$ values and complexity scores (Algorithm \ref{alg:nsga}, Line 4).
During pipeline evaluation, a pipeline is trained on the training set and its $r^2$ is determined on the validation set.
Upon completing a pipeline evaluation, we record the pipeline's $r^2$ value, complexity score, and set of SNPs removed during the LD pruning step (Section \ref{sec:ld_prune_node}).
Pipelines that fail during evaluation--such as those that do not retain any SNPs after both feature selection steps--are removed from the population.
Additionally, pipelines that receive a negative $r^2$ value are also discarded from the population.
After evaluating pipelines, each LD-pruned SNP is sent to update SNP DB (Algorithm \ref{alg:nsga}, Line 5).

The initial population enters the evolutionary loop after all pipelines receive an $r^2$ and complexity score.
The first step in this loop is to select a sufficient number of parent pipelines with binary nondominated tournament selection (Algorithm \ref{alg:nsga}, Line 7), as is standard in NSGA-II \cite{deb2002nsgaii}.
Before selecting a parent, all pipelines are assigned a Pareto front ranking and a crowding distance derived from their $r^2$ and complexity scores.
Here, pipeline A dominates pipeline B if:
\begin{align}
    A_{r^2} \geq B_{r^2} &\text{ and } A_{complexity} \leq B_{complexity} \\
    A_{r^2} > B_{r^2} &\text{ or } A_{complexity} < B_{complexity}
\end{align}

Once all pipelines have been assigned a front ranking and crowding distance, parent selection can be performed where lower front rankings and higher crowding distances are preferred.
The set of parents provides the genetic material necessary to generate the set of offspring pipelines (Algorithm \ref{alg:nsga}, Line 8).
While NSGA-II produces an offspring set equal in size to the initial population, in this work, we instead generate twice as many offspring.
Specific details concerning the generation of offspring are provided in Section \ref{sec:offspring_generation}.

Offspring follow a similar process as their parents: 
update the SNP DB with newly introduced SNPs (Algorithm \ref{alg:nsga}, Line 9), 
compute $r^2$ and complexity score (Algorithm \ref{alg:nsga}, Line 10), 
and process all LD-pruned SNPs (Algorithm \ref{alg:nsga}, Line 11).
The offspring set is reduced to the initial population $size$ via survival selection (Algorithm \ref{alg:nsga}, Line 12), prioritizing offspring with lower front ranks and higher crowding distances.
Before survival selection, however, we eliminate pipelines that exhibit identical complexity scores resulting from the same set of SNPs being sent to a regressor. 
Among these redundant pipelines, one is randomly selected to proceed.
It is important to note that only offspring undergo survival selection, diverging from the original NSGA-II implementation.
If the number of offspring passed to survival selection is lower than the initial population $size$, all offspring survive.

The surviving offspring form the new population and undergo the same cycle as their predecessors, continuing this process until the maximum number of generations is reached (Algorithm \ref{alg:nsga}, Lines 8 - 14).
We deviate from the offspring generation and survival selection of NSGA-II to maintain pipelines with relevant SNPs, as parent pipelines may retain suboptimal/pruned SNPs.
At the end of the run, we return the final population for further analysis (Algorithm \ref{alg:nsga}, Line 14).

\subsection{Offspring Generation}
\label{sec:offspring_generation}
Offspring pipelines are generated using GP (Section \ref{sec:engine}) enhanced with knowledge extracted from the SNP DB (Section \ref{sec:snp:data}), thereby reducing evaluation of suboptimal pipelines.
The GP framework builds offspring via crossover and mutation, and the SNP DB enhances this process by recommending SNPs to incorporate within an offspring.
These recommendations may be \textit{smart}, prioritizing SNPs based on marginal validation $r^2$, or \textit{random} by recommending SNPs without bias.

The probability of offspring generated through mutation or crossover is specified by the user.
Regardless of how offspring are built, their pipeline structure is inherited from their parents.
Specifically, a parent's LD-pruning, feature selection, and regressor nodes are integrated within an offspring.
However, not all SNPs present in the parent are transmitted; only those that persist after LD-pruning and feature selection are considered.
This mechanism ensures that the offspring start with a subset of SNPs that contributed to the parent’s success within the population, serving as promising building blocks for further optimization.
The offspring is considered fully constructed once crossover and mutation operations are applied.

\subsubsection{Offspring Via Mutation}
\label{sec:mutation}
Offspring generated via mutation are derived from a single parent.
Mutations are applied with equal probability to the LD-pruning, feature selection, or regressor nodes, with shifts introduced to their associated hyperparameters. 
The initial set of SNPs in the offspring is not directly mutated; rather, each SNP serves as an anchor for identifying additional SNPs to extend the current set of SNPs.
The count of SNPs to mutate is sampled from the minimum and maximum constraints, which is then used to sample, with replacement, a set of SNPs to mutate from the initial parent set.
Details on SNP mutation are in Section \ref{sec:snp:recs}.

\subsubsection{Offspring via crossover}
Offspring generated via crossover are derived from two parents.
With equal probability, the offspring inherits the LD-pruning, feature selection, and regressor nodes from either parent. 
The inheritable SNPs are grouped and then sampled to form the subset inherited by the offspring. 
The \textit{count} of SNPs to sample is drawn from the minimum and maximum SNP constraints.
Under a smart approach, SNPs are prioritized based on their marginal $r^2$ (Section \ref{sec:snp:data}), whereas a random approach treats all SNPs equally.
If the number of inheritable SNPs is less than the sampled count, all SNPs are included.
This mechanism promotes the propagation of informative SNPs while allowing the emergence of pipelines with a small SNP \textit{count}.
Additionally, there is a chance (determined by the user-specified mutation probability) that the offspring generated through crossover also undergoes mutation (Section \ref{sec:mutation}).
However, all SNPs assigned to offspring are eligible for mutation.

\subsubsection{SNP node mutation}
\label{sec:snp:recs}
When a SNP undergoes mutation, SNP DB recommends new SNPs to append to offspring. 
This mutation process does not alter SNPs directly.
One of three distinct mutation strategies with equal probability are applied:
\begin{itemize}
    \item \textbf{Within Bin}: Select a SNP from the same bin.
    \item \textbf{Within Chromosome}: Select a SNP within the same chromosome, excluding those in the same bin.
    \item \textbf{Outside Chromosome}: Select a SNP from a different chromosome.
\end{itemize}
These strategies generate SNP recommendations with varying degrees of locality: 
\textbf{Within Bin} supports the exploitation of specific genomic regions, 
\textbf{Within Chromosome} enables broader exploration in a single chromosome to identify potentially superior peaks, 
and \textbf{Outside Chromosome} encourages global exploration across the entire genome.

The three mutation strategies support \textit{smart} and \textit{random} sampling.
In smart sampling, only SNPs that have a marginal validation $r^2$ and a \textit{true} consideration flag are eligible (Section \ref{sec:snp:data}). 
These SNPs are weighted by their marginal validation $r^2$ (Section \ref{sec:snp:data}).
In contrast, random sampling considers all SNPs with a \textit{true} consideration flag. 
If the smart mutation fails to return a SNP, we default to a random mutation.
Each approach is allowed up to 20 attempts to select a SNP. 
If both strategies fail, a fallback mechanism selects a random SNP from the entire set of SNPs with \textit{true} consideration flags.

\subsection{SNP consistency score}
\label{sec:pfi}
We extract the Pareto-optimal pipelines from the final population to compute feature importance for each SNP present in these pipelines.
However, this analysis only considers SNPs that contribute to pipeline complexity and have a \textit{true} consideration flag.
StarBASE-GP uses permutation feature importance (PFI) \cite{altmann2010permutation, breiman2001random} to quantify the contribution of each SNP to model performance.
PFI is a model-agnostic technique that assesses feature importance by permuting the values of each predictive feature (SNP) and measuring the resulting change in model performance.
A substantial reduction in performance upon permutation indicates a strong dependence between the feature and the target, signifying high importance; conversely, a negligible or opposite impact suggests lower importance.
Specifically, we use the scikit-learn implementation of PFI with $100$ permutations for each SNP.

Once PFI scores have been assigned to all SNPs within each pipeline, we compute the \textit{mean rank} and \textit{appearance proportion} for each unique SNP represented in the Pareto front.
Within each pipeline, SNPs are ranked based on their PFI in descending order, with the SNP with the highest PFI ranked as $1$.
The mean rank of each unique Pareto SNP is calculated by averaging its ranks across the pipelines in which it appears.
A lower mean rank implies that the SNP is a key feature contributing significantly to predictive performance, as it is prioritized within individual pipelines. 
The appearance proportion is the proportion of Pareto pipelines in which the SNP is present.
A higher appearance proportion indicates that a SNP is consistently selected across multiple Pareto-optimal pipelines, suggesting its potential biological relevance. 
With these two values, we compute a \textit{SNP Consistency Score} for each unique SNP represented in the Pareto front, given by:
$$\frac{1}{\text{Mean Rank}} \times \text{Appearance Proportion}$$
Together, mean rank and appearance proportion are integrated into a novel composite score that identifies SNPs that are both impactful and consistently found across pipelines, with values ranging within the interval $(0.0, 1.0]$.
\section{Methods}
To evaluate StarBASE-GP, we apply it to a real-world, publicly available dataset (Section \ref{sec:data}).
StarBASE-GP operates under a fixed budget of 30,150 pipeline evaluations, with its configuration outlined in Table \ref{tab:gp_parameters}.
We evaluate its performance against two additional experimental conditions: (1) a random control, representing unguided exploration of pipeline configurations, and (2) a basic GP approach that does not incorporate biological domain knowledge. 
The random control serves as a baseline for evaluating pipeline building in the absence of heuristic guidance.
In contrast, the basic GP system isolates the impact of incorporating biological knowledge in the evolutionary optimization of SNP subsets. 
Both conditions are allocated the same evaluation budget, and we conduct 40 replicates for each experimental condition.

\subsection{Additional experimental conditions}
\subsubsection{Random control}
This condition resembles how the population is initialized in StarBASE-GP (Section \ref{sec:engine}), with the caveat that the population is initialized with 30,150 pipelines.
After all pipelines are evaluated, we remove those that failed during evaluation. % and produced a negative pipeline $r^2$.
We conduct the same analysis with the remaining pipelines described in Section \ref{sec:pfi}.

\subsubsection{Basic GP}
This condition deviates from StarBASE-GP in the following ways: the pipelines omit the LD-pruning node, all SNPs retain a \textit{true} consideration flag within the SNP DB, and no intelligent sampling strategies are used for SNP recommendation. 
Offspring are generated using the same procedures as in StarBASE-GP; however, only randomly selected SNPs are recommended for mutation, and SNPs are sampled randomly for inclusion during crossover. 
This setup follows the applicable GP configurations detailed in Table \ref{tab:gp_parameters}.

\subsection{Data}
\label{sec:data}
The dataset consists of 128,400 SNPs from an outbred, related brown rat (\textit{Rattus norvegicus}) population consisting of males and females derived from eight inbred founders (Heterogeneous Stock\cite{hansen1984development}).
This dataset is chosen due to its controlled breeding design, which reduces confounding genetic and environmental factors commonly encountered in human studies.
These SNPs (Rnor 6.0 assembly) were previously used in two GWAS investigating obesity-related traits in \textit{R. norvegicus} \cite{chitre2020genome, freda2024pager}.
For the phenotype, we use quantile-normalized residuals of body mass index including the rat’s tail (BMI\_Tail) due to its large sample size (\textit{n} = $3,166$), relatively high heritability ($0.31$ ± $0.03$)\cite{chitre2020genome}, and for comparison to four QTLs identified in the prior studies\cite{chitre2020genome, freda2024pager}.

We use a $50$/$50$ split in all three experiments to maximize representation of data samples in both the training and validation sets.
Thus, the maximum number of SNPs per pipeline is set to $150$ (Table \ref{tab:gp_parameters}), based on the calculation: $3,166 \div 2 \times 0.10 = 158.3$, which we round down to $150$ as a conservative and convenient setting that remains within the bounds of the ``$10x$ rule" (Section \ref{sec:SNP_node}).
We set the minimum number of SNPs per pipeline to $50$ to ensure the SNP subsets generated via mutation are sufficiently informative while discouraging large feature sets.

The original researchers normalized BMI\_Tail to control for batch and sex effects \cite{chitre2020genome}. 
We further adjust BMI\_Tail by performing a principal component analysis using the genotype data in PLINK2 \cite{purcell2007plink} within R \cite{Rcitation}. 
To do so, we perform a multiple linear regression of the original normalized phenotype on the first ten principal components, retaining the residuals as the new phenotype for all analyses.
Since StarBASE-GP requires genotype data to be encoded as $0$, $1$, or $2$ (additive encoding), we use this approach instead of incorporating a genetic relatedness matrix, which is commonly used in GWAS models to control for biases introduced by high relatedness \cite{uffelmann2021genome}.
Here, a similar effect is achieved without modifying the underlying genotype encoding\cite{price2006principal}.

\begin{table}[h]
\centering
\footnotesize
\caption{GP Configuration. Bolded values are used by both GP systems.}
\vspace{0em}
\begin{tabular}{|l|l|}
\hline
\multicolumn{2}{|l|}{\textbf{\textit{GP Parameters}}} \\
\hline
\textbf{Population Size} & 150 \\
\textbf{Number of Generations} & 100 \\
\textbf{Minimum SNPs per Pipeline} & 50 \\
\textbf{Maximum SNPs per Pipeline} & 150 \\
Bin Size & 500 \\
\hline
\multicolumn{2}{|l|}{\textbf{\textit{Variation Operator Probability}}} \\
\hline
\textbf{Crossover}  & 0.50 \\
\textbf{Mutation}  & 0.50 \\
\hspace{3mm}Within Bin  & 0.33 \\
\hspace{3mm}Within Chromosome  & 0.33 \\
\hspace{3mm}Outside Chromosome  & 0.33 \\
Random Strategy  & 0.75 \\
Smart Strategy  & 0.25 \\
LD, \textbf{Selector}, \textbf{Regressor} \textbf{Tuning}  & 0.50 \\
\hline
\end{tabular}
\label{tab:gp_parameters}
\end{table}

The 128,400 SNPs are the result of an LD-pruning process applied to an initial set of over $3.4$ million SNPs\cite{chitre2020genome}.
However, a liberal threshold of \(r^2 = 0.95\) was used.
Thus, StarBASE-GP’s LD operator can reduce the effects of LD more rigorously.
Finally, we filter out SNPs with zero variance ($3$), $\geq$ $5\%$ missingness (17,916), or a minor allele frequency (MAF) $\leq$ $0.01$ (6,667), resulting in a final dataset of 103,814 SNPs.
Remaining missing genotypes ($1.15\%$) are mode-imputed per locus to ensure data completeness.

\subsection{Data tracking and statistical testing}
Each replicate, across all experimental conditions, generates a set of summary files. 
These include details on Pareto fronts, data coverage, runtime, SNP DB content, QTL identification accuracy, and per-SNP metrics including optimal encoders, and feature importance.
All code used for the analysis, visualization, and statistical testing of these summary files is found in our supplemental information.
All statistical tests are conducted with a significance level of $\alpha = 0.05$ and we apply FDR correction for multiple tests.

\subsubsection{Data coverage and runtime}
For each experiment, we calculate mean data coverage and apply a Kruskal-Wallis test to assess overall differences across experiments. 
We then conduct one-tailed Wilcoxon rank-sum tests under the directional hypothesis that data missingness is lower in random controls compared to both basic GP and StarBASE-GP, and lower in StarBASE-GP compared to basic GP. 
We repeat this procedure for mean computation time (hours to finish all generations); however, the directional hypothesis here is that computation time is greater in random compared to the other experiments, and greater in basic GP compared to StarBASE-GP.

\subsubsection{Pareto front analysis}
We analyze three characteristics of the Pareto fronts generated from the final population under each experimental condition: hypervolumne, the number of non-dominated solutions, and SNP diversity.  
Hypervolume \cite{zitzler1999evolutionary, zitzler2000comparison} is calculated as the volume of the objective space dominated by the Pareto front.
For this calculation, pipeline $r^2$ is used directly.
In contrast, pipeline complexity is transformed into a maximization objective: the raw complexity is first normalized by subtracting the minimum possible value ($1$) and dividing by the range ($149$), corresponding to the difference between the maximum ($150$) and minimum ($1$) possible SNP counts.
This transformed value is then subtracted from $1$ to reflect the inverse relationship between complexity and optimality.
SNP diversity is measured as the number of unique bins represented in each replicate's Pareto front.

We apply a Kruskal-Wallis test to assess whether significant differences occur among all experimental conditions for each Pareto front characteristic.
We then conduct a one-tailed Wilcoxon rank-sum test under the directional hypothesis that the hypervolume and size of the Pareto front are lower in random controls than in basic GP and StarBASE-GP, and lower in basic GP than in StarBASE-GP.
We test under the directional hypothesis that SNP diversity is greater in random controls compared to the other experiments, and lower in basic GP compared to StarBASE-GP.

\subsubsection{Accuracy and precision}
Accuracy and precision are assessed based on each experiment’s ability to identify four QTLs previously reported in two GWAS studies\cite{chitre2020genome, freda2024pager}. 
However, because these prior studies did not consider all nine inheritance models used here (Chitre \textit{et al.}\cite{chitre2020genome}: additive; Freda \textit{et al.}\cite{freda2024pager}: additive, recessive, dominant, PAGER), the chromosomal position and optimal inheritance model for each QTL may differ, as each locus can yield higher or lower marginal $r^2$ depending on the encoding applied.
To define ground truth targets, we isolate all SNPs on QTL chromosomes reported in the original studies.
We then regress the BMI\_Tail phenotype on each SNP using all nine inheritance models across the full sample ($n = 3,166$).
For each QTL region, the SNP and inheritance model combination with the highest marginal $r^2$ is designated as the target (Table \ref{tab:QTL_inhertiance}).

\begin{table}[ht]
\centering
\caption{BMI\_Tail QTL Positions and Inheritance Models (A: Additive, P: PAGER, R: Recessive, SA: Subadditive, D: Dominant).}
\vspace{0em}
\footnotesize
\begin{tabular}{|c|c|c|c|}
\hline
\textbf{QTL} & \textbf{Chitre et al.\cite{chitre2020genome}} & \textbf{Freda et al.\cite{freda2024pager}} & \textbf{StarBASE-GP} \\
\hline
1 & \makecell[l]{1.106866154 (A)} & \makecell[l]{---} & \makecell[l]{1.106956497 (A)} \\
2 & \makecell[l]{1.282049439 (A)} & \makecell[l]{1.281489331 (P)} & \makecell[l]{1.281489331 (SA)} \\
3 & \makecell[l]{10.84080794 (A)} & \makecell[l]{10.84021443 (P)} & \makecell[l]{10.84021443 (A)} \\
4 & \makecell[l]{---} & \makecell[l]{18.27355039 (R)} & \makecell[l]{18.27355039 (D)} \\
\hline
\end{tabular}
\label{tab:QTL_inhertiance}
\end{table}

We assess \textit{QTL accuracy} as the count of the four known QTLs that are contained in each replicate’s final Pareto front, where a QTL is considered identified if any SNP within ±$1$ million base pairs ($1$Mb) of the target locus is present.
This window accounts for variation in peak location across data splits, as marginal $r^2$ values can shift depending on the sample. 
\textit{Overall QTL accuracy} is reported as the summation of QTL replicate accuracy divided by the maximum possible number of QTL hits ($160$ total).
\textit{QTL precision} quantifies how closely Pareto SNPs in each replicate align with QTL genomic locations (Table \ref{tab:QTL_inhertiance}), calculated as the relative bp distance between peaks and respective targets within a ±$1$Mb window. 

To compare QTL accuracy across experiments, we use a Kruskal-Wallis test followed by one-tailed Brunner-Munzel tests under the directional hypothesis that StarBASE-GP outperforms basic GP and random controls, and that basic GP outperforms random controls. 
For QTL precision, we apply a Kruskal-Wallis test followed by one-tailed Wilcoxon rank-sum tests, based on the directional hypothesis that random controls exhibit greater distances from target QTLs than basic GP and StarBASE-GP, and that basic GP distances are greater than StarBASE-GP's. 
Tests for QTL precision were conducted both per QTL and across all QTLs.

\subsubsection{SNP consistency score evaluation and comparison}
For each experimental condition, we compute the mean SNP consistency score for every SNP that appears in the Pareto front of at least two replicates.
We then visualize these scores using Manhattan plots to highlight peaks of phenotypic association.
To compare the most explanatory SNPs across experiments, we select the top $50$ SNPs based on their mean SNP consistency score across replicates, accounting for variation in peak locations introduced by data splitting.
For each ±$1$Mb window within this top-ranked set, we identify the most frequently replicated SNP, record its mean consistency score and most frequently selected encoder, and designate that SNP as the representative peak for the genomic region.
This approach prioritizes SNPs that are both stable across replicates and strongly associated with the phenotype, allowing us to evaluate recovery of known QTLs and identify potentially novel loci.

\subsubsection{Inheritance model validation}
To assess how well StarBASE-GP models SNP-level genetic architecture, we compare observed phenotype distributions across genotype classes in the full dataset for SNPs of interest to eight strict inheritance models (Figure \ref{fig:inheritance}) and their mirrored versions. 
Each model is scaled to the range of BMI\_TAIL residuals, and the best-fit model is identified by the highest Pearson correlation between observed and expected phenotypic values. 
We then compare these best-fit models to the most frequently selected encoders assigned by StarBASE-GP for each SNP of interest.
\section{Results \& Discussion}
\subsection{StarBASE-GP balances coverage and runtime efficiency}
\label{sec:res:coverage_runtime}
All three experiments exhibit minimal data missingness (below $0.15\%$), with the random control showing significantly less missingness than both GP-based methods, and StarBASE-GP outperforming basic GP (Table \ref{tab:compute_coverage}; Supplementary File S1).
High data coverage is critical to ensure that the vast majority of SNPs are evaluated for phenotypic relevance.

\begin{table}[ht]
\centering
\caption{Mean (± S.E.) runtime and data coverage by experiment.}
\vspace{0em}
\footnotesize
\begin{tabular}{|c|c|c|}
\hline
\textbf{Experiment} & \textbf{Data Coverage (\%)} & \textbf{Runtime (h)} \\
\hline
Random & $99.99 \pm 2.41 \times 10^{-5}$ & $5.46 \pm 0.04$ \\
Basic GP & $99.86 \pm 0.01$ & $8.83 \pm 1.12$ \\
StarBASE-GP & $99.92 \pm 0.02$ & $5.45 \pm 0.31$ \\
\hline
\end{tabular}
\label{tab:compute_coverage}
\end{table}

Although the random control and StarBASE-GP have similar mean runtimes (\textasciitilde$5.5$ hours; Table \ref{tab:compute_coverage}), random’s runtime is significantly longer, likely due to lower variability across replicates. 
Although not significantly more than the other experiments, basic GP has a higher mean runtime of \textasciitilde$9$ hours.
In general, lower runtime is desirable as long as results are not compromised.
In this regard, StarBASE-GP offers an efficient balance between data coverage and computational cost.

\subsection{StarBASE-GP yields superior Pareto front characteristics}
\subsubsection{Hypervolume}
\label{sec:res:hypervolume}

\begin{figure}[!h]
\centering
\includegraphics[width=\linewidth]{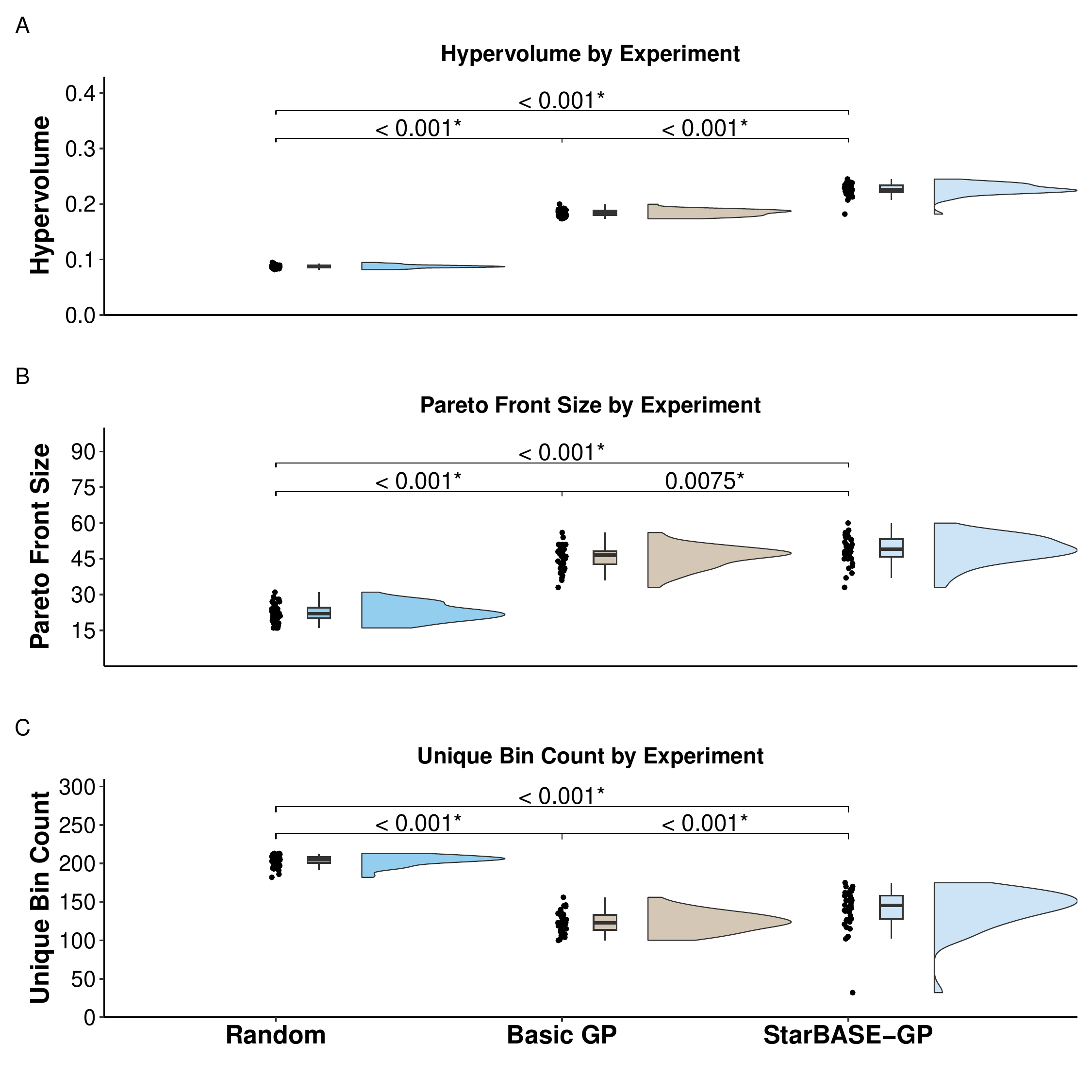}
\caption{Characteristics of the Pareto fronts for each experimental condition. Panels present raincloud plots for hypervolume (A), number of front solutions (B), and the count of unique bins (C).}
\label{fig:pareto-characteristics}
\end{figure}

StarBASE-GP produces fronts with significantly greater hypervolume compared to both other experimental conditions (Figure \ref{fig:pareto-characteristics}A).
Although the basic GP condition achieves perfect overall QTL accuracy (Table \ref{tab:qtl_accuracy}), it fails to produce fronts with hypervolumes similar to those of StarBASE-GP. 
The sole distinction between StarBASE-GP and basic GP lies in the incorporation of biological knowledge. % during the evolutionary search. 
Specifically, StarBASE-GP leverages the SNP DB to guide pipeline development and reduce the search space--mechanisms absent in basic GP. 
In contrast, the random control lacks any mechanism for pipeline improvement. 
Consequently, it is unsurprising that this condition yields significantly lower hypervolumes relative to both other conditions.

In the context of GPA tasks, attaining perfect explainability (pipeline $r^2$ of $1.0$) is unrealistic.
This limitation stems from the fact that pipelines may be working with data that omits features necessary to fully explain a phenotype (Section \ref{sec:intro}).
Nonetheless, hypervolume remains an informative metric for assessing the quality of Pareto fronts.
In the ideal Pareto front, a pipeline with the lowest complexity includes the most informative SNP, maximizing $r^2$. 
Pipelines with increasing complexity should incorporate this SNP, along with the next most informative SNP, and so on.
Thus, larger hypervolume, like that produced by StarBASE-GP, indicates progressive inclusion of informative SNPs driven by biological knowledge.

\subsubsection{Pareto front pipeline count}
\label{sec:res:front_size}
The number of Pareto optimal pipelines within the final population reflects the number of distinct sets of SNPs maintained.
StarBASE-GP produces Pareto fronts with significantly more pipelines compared to both other experimental conditions (Figure \ref{fig:pareto-characteristics}B).
The pipeline count provides insights into the balance between exploitation and exploration achieved under each experimental condition.
A smaller front may indicate a strategy that optimizes a limited number of pipelines or has converged to local optima. 
In contrast, a larger front suggests that a strategy better explores a pipeline space, yielding a more diverse set of pipelines with various tradeoffs.
However, filling these positions depends on how effectively an approach optimizes pipelines.

The random control produces significantly smaller Pareto fronts than both other experimental conditions using GP.
Both GP systems initialize a population of randomly generated pipelines; however, we hypothesize that StarBASE-GP’s integration of biological knowledge during offspring construction contributes to larger fronts. 
Specifically, offspring produced via mutation in StarBASE-GP may extend the front around the parent by introducing informative SNPs suggested by the SNP DB.
Additionally, offspring generated through crossover with smart sampling inherit promising SNPs from different regions of the front, further promoting diversity and expansion.
For instance, when crossover occurs between two parents positioned at opposite ends of the complexity spectrum, the resulting offspring may inherit key SNPs identified in the lower-complexity parent, while also integrating novel variants from the higher-complexity counterpart.
In contrast, offspring produced by basic GP through mutation or crossover incorporate SNPs selected at random, which does not guarantee the inclusion of informative building blocks.

In the context of the SNP Consistency Score, the size and quality of the Pareto front directly influence both the mean rank and appearance proportion. 
For instance, fronts of identical size but differing hypervolumes indicate distinct SNP subsets.
The front exhibiting higher hypervolume likely contains more relevant SNPs with higher SNP consistency scores, which is reflected in the fronts generated by StarBASE-GP.
These results, together with the hypervolume findings presented in Section \ref{sec:res:hypervolume}, provide strong evidence that StarBASE-GP is capable of exploring, optimizing, and propagating multiple SNP combinations, thereby improving pipeline $r^2$ for a given set size.

\subsubsection{SNP diversity}
\label{sec:res:snp_dei}
We measure SNP diversity by counting the number of unique bins represented across all SNPs within a given Pareto front (Figure \ref{fig:pareto-characteristics}C), where each bin corresponds to a genomic region, thereby reflecting coverage across the entire genome.
The random control yields fronts with a significantly higher number of bins than both other experimental conditions.
This outcome is expected, as the random control assembles all pipelines with random SNPs, enabling diverse SNP combinations on the front by chance.

In contrast, basic GP produces Pareto fronts with a significantly lower number of bins compared to both other experimental conditions. 
Basic GP lacks a mechanism to recommend informative SNPs for mutation and crossover.
Consequently, pipelines that acquire informative SNPs are incentivized to retain and propagate these SNPs extensively to maintain their lineage across the front, which occurs when offspring are generated via mutation.
Moreover, there is selection pressure to incorporate neighboring SNPs around peaks as offspring generated via crossover cannot preferentially incorporate the most informative SNPs; thus, increasing the density of neighboring informative SNPs in pipelines enhances the likelihood that these SNPs are transmitted to offspring. 
As a result, basic GP applies selection pressure favoring the inclusion of SNPs near informative peaks, a behavior that aligns with the observed reduction in SNP diversity across its fronts.

The ability to identify the most informative SNPs for a given GPA task depends on two factors: (1) comprehensive coverage of SNPs in a dataset, and (2) constructing pipelines with informative SNP subsets. 
As discussed in Section \ref{sec:res:coverage_runtime}, all three experimental conditions achieve over $99\%$ coverage.
The remaining challenge lies in assembling SNP combinations within pipelines that maximize $r^2$, where the promotion of SNP diversity plays a critical role.
For example, prioritizing diversity alone may omit key building block SNPs, as the emphasis on dissimilarity can overlook performance.
Conversely, neglecting diversity may result in pipelines comprising redundant or highly correlated SNPs, thereby increasing the risk of premature convergence to local optima.
Thus, there must be a balance between the exploration of diverse SNP subsets and the exploitation of promising candidates to consistently identify the most informative variants.
StarBASE-GP achieves an intermediate level of diversity compared to the other experimental conditions, supporting this balance.

\subsection{GP-based searches have higher accuracies and precisions compared to random search}
\label{sec:res:accuracy}
StarBASE-GP and basic GP achieve overall QTL accuracies of $0.99$ and $1.00$, respectively. 
In contrast, the random control is significantly less accurate, at $0.76$ (Table \ref{tab:qtl_accuracy}; Supplementary File S1). 
In one replicate (Seed $251$), StarBASE-GP fails to include QTL $10.84021443$ in the final Pareto front, despite identifying it during the evolutionary run (Supplemental Information).
In this replicate, signal fluctuations from data splitting result in a negative marginal validation $r^2$ for $10.84021443$, leading to it being pruned. 
SNP $10.93564208$ emerges as the regional peak; however, because it lies $>1$ Mb away, it is not counted as a QTL hit for this locus.

\begin{table}[ht]
\centering
\caption{QTL accuracy and overall accuracy by experiment.}
\vspace{0em}
\footnotesize
\begin{tabular}{|c|c|c|c|c|c|c|}
\hline
\multirow{2}{*}{\textbf{Experiment}} & \multicolumn{5}{c|}{\textbf{QTL Hits}} & \multirow{2}{*}{\textbf{Accuracy}} \\
\cline{2-6}
& \textbf{0} & \textbf{1} & \textbf{2} & \textbf{3} & \textbf{4} & \\
\hline
Random & 0 & 2 & 9 & 15 & 14 & 0.76 \\
Basic GP & 0 & 0 & 0 & 0 & 40 & 1.00 \\
StarBASE-GP & 0 & 0 & 0 & 1 & 39 & 0.99 \\
\hline
\end{tabular}
\label{tab:qtl_accuracy}
\end{table}

Although StarBASE-GP may occasionally exclude a true QTL due to local fluctuations in marginal $r^2$, this reflects a deliberate design choice: by selecting only the most explanatory SNP within a given genomic region, StarBASE-GP avoids redundancy and allocates front space to distinct, high-confidence signals.
Over multiple replicates, this strategy enables StarBASE-GP to consistently recover true genomic peaks, while basic GP--lacking pruning mechanisms--tends to fill Pareto fronts with redundant SNPs that echo the same underlying signal (see Section \ref{sec:res:snp_dei}).
The significantly lower accuracy we observe in the random control highlights the importance of an evolutionary strategy in the GPA problem domain.
Since the random control relies entirely on chance to populate the Pareto front, it is unlikely to consistently capture all informative SNPs in pipelines given the vast search space.

\begin{table}[ht]
\centering
\caption{QTL precision (in kilobases, rounded to nearest kb ± S.E.) across experiments.}
\vspace{0em}
\footnotesize
\begin{tabular}{|c|c|c|c|}
\hline
\textbf{QTL} & \textbf{Random} & \textbf{Basic GP} & \textbf{StarBASE-GP} \\
\hline
1.106956497 & $301 \pm 47$ & $176 \pm 40$ & $85 \pm 33$ \\
1.281489331 & $319 \pm 45$ & $238 \pm 40$ & $175 \pm 41$ \\
10.84021443 & $322 \pm 76$ & $76 \pm 24$ & $87 \pm 26$ \\
18.27355039 & $315 \pm 57$ & $46 \pm 14$ & $37 \pm 20$ \\
\hline
\textbf{All QTLs} & \textbf{313 ± 27} & \textbf{134 ± 17} & \textbf{96 ± 16} \\
\hline
\end{tabular}
\label{tab:qtl_positional_accuracy}
\end{table}

Except for $10.84021443$, StarBASE-GP consistently achieves higher QTL precision than the other conditions (Table \ref{tab:qtl_positional_accuracy}). 
Across all QTLs, the genomic distance between Pareto SNPs and the true targets is significantly lower in StarBASE-GP compared to alternative methods (Supplementary File S1). 
These results highlight the critical role of the LD operator in GPA tasks. 
By pruning redundant SNPs surrounding peaks of phenotypic assocation, StarBASE-GP effectively “hillclimbs” toward the true causal signals with greater precision. 
This confers two major advantages: (1) it reduces the effective search space, and (2) it frees capacity to explore additional, potentially informative regions of the genome.

While basic GP achieves better precision at $10.84021443$, as mentioned previously, fluctuations in marginal $r^2$ values caused by data splitting can occasionally shift peak identification in some replicates. 
Compounding this, the genomic region surrounding $10.84021443$ is relatively sparse in SNP density (Supplementary File S1), increasing the likelihood of distant alternative peaks being selected. 
Nevertheless, across replicates, StarBASE-GP consistently converges on $10.84021443$ as the peak for this region (Table \ref{tab:qtl_accuracy}).

\subsection{StarBASE-GP identifies ground truth and novel QTLs with robust SNP consistency scores}
\label{sec:res:SNP_consistency}

\begin{figure}[!h]
\centering
\includegraphics[width=\linewidth]{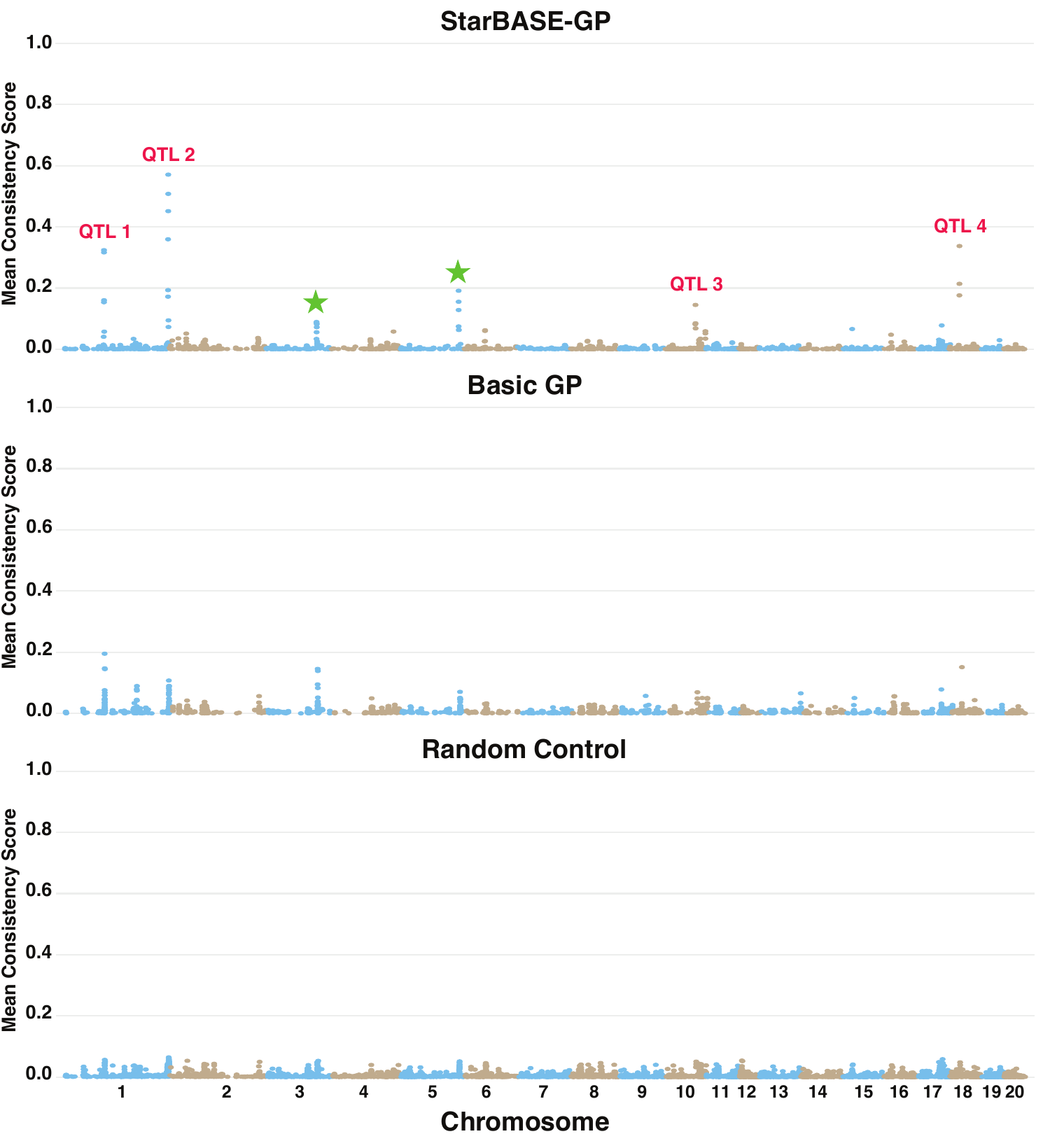}
\caption{Manhattan plots of mean SNP consistency scores across replicates for Pareto front SNPs by experiment. Red "QTL" labels mark genomic locations of the four ground truth QTLs. Green stars denote potentially novel QTLs.}
\label{fig:manhattan-SNPconsistency}
\end{figure}

StarBASE-GP outperforms both basic GP and the random control in SNP consistency scores at all four ground truth QTLs (Figure \ref{fig:manhattan-SNPconsistency}; Supplementary File S1). 
StarBASE-GP exhibits clear, well-defined signal peaks, whereas the signals from the other methods are generally weak or diffuse. 
The low signals observed in basic GP and the random control are likely due to their inability to consistently assemble informative SNP combinations within pipelines. 
In the random control, pipelines are generated without selection pressure or refinement, making it improbable that informative building blocks co-occur with sufficient frequency to yield strong signals. 
In basic GP, although evolution is present, the absence of biologically guided SNP selection results in pipelines populated by redundant or non-informative variants. 
These issues dilute individual feature contributions, reducing overall SNP consistency scores and masking true signals.

A noteworthy difference between StarBASE-GP and basic GP is that basic GP ranks QTL 1 as the most informative locus, followed by QTL 4 (Figure \ref{fig:manhattan-SNPconsistency}). 
In contrast, StarBASE-GP’s rank ordering aligns with prior findings, with QTL 2 showing the strongest phenotypic association\cite{chitre2020genome, freda2024pager}, followed by QTL 4\cite{freda2024pager}. 
Although the previous studies found similar signal strengths for QTLs 1 and 3\cite{chitre2020genome, freda2024pager}, both StarBASE-GP and basic GP identify a stronger signal at QTL 1. 
This discrepancy may be driven not only by increased SNP density in the QTL 1 region and the enhanced encoding flexibility within our framework, but also by fundamental differences between feature importance scores and traditional \textit{p}-values. 
Unlike \textit{p}-values derived from GWAS, SNP consistency scores reflect predictive performance and stability across replicates, which may lead to shifts in signal strength and prioritization.
Nevertheless, the consistency of StarBASE-GP’s results across replication underscores the predictive reliability of these loci and offers complementary insights into phenotypic association.

It is particularly notable that StarBASE-GP identifies all four QTLs as highly informative peaks, as neither of the previous GWAS studies detected the full set.  
Chitre \textit{et al.} (2020) failed to detect QTL 4 due to the exclusive use of an additive encoding\cite{chitre2020genome}, while Freda \textit{et al.} (2024) failed to detect QTL 1 as a result of the strict significance threshold applied\cite{freda2024pager}.  
This highlights StarBASE-GP’s ability to recover true signals that may be overlooked by conventional modeling or thresholding strategies.

StarBASE-GP identifies two potentially novel QTLs on chromosome 3 ($3.136975356$) and chromosome 5 ($5.161222644$) (Figure \ref{fig:manhattan-SNPconsistency}).
Notably, basic GP also detects a peak on chromosome 3 at a nearby position ($3.136971250$).
The chromosome 3 locus lies within the same genomic region as pleiotropic QTLs reported by Chitre \textit{et al.} (2020) at positions $3.136021511$ and $3.137537161$, which were associated with body length (with and without tail), body weight, and retroperitoneal fat under an additive model\cite{chitre2020genome}.
These traits are closely related to BMI, and StarBASE-GP’s ability to identify this locus—despite its failure to surpass conventional GWAS significance thresholds for BMI\_TAIL in prior studies\cite{chitre2020genome, freda2024pager}—underscores its sensitivity to biologically relevant loci.
Although this locus ranks lower in SNP consistency compared to ground truth QTLs, its biological plausibility and consistency across experiments highlight it as a promising candidate for further investigation.

The chromosome 5 locus was not implicated in either of the previous studies\cite{chitre2020genome, freda2024pager}; however, in StarBASE-GP, it exhibits a stronger signal than the chromosome 3 locus and even surpasses that of QTL 3 (Figure \ref{fig:manhattan-SNPconsistency}).
This SNP resides within a large intron of the \textit{Kazn} (kazrin, periplakin-interacting protein) gene in \textit{R. norvegicus}\cite{vedi20232022}.
Although introns are not translated into protein, they can play crucial regulatory roles in gene expression and alternative splicing\cite{shaul2017introns}.
\textit{Kazn} is predicted to be involved in keratinization—a process increasingly linked to obesity through inflammation and pleiotropic connections with metabolic and endocrine function\cite{ellulu2017obesity}.
Furthermore, \textit{Kazn} has been identified as an integral component of desmosomes\cite{groot2004kazrin}, which are critical for maintaining epithelial integrity and cellular adhesion under mechanical and inflammatory stress\cite{green2007desmosomes}.

Taken together, these findings demonstrate how StarBASE-GP, unbound by conventional significance tests and thresholds, can prioritize novel loci overlooked by traditional approaches.
This capacity supports downstream validation and offers valuable paths forward for the discovery of novel contributors to complex traits and disease susceptibility.

\subsection{StarBASE-GP accurately models inheritance patterns of informative SNPs}
\label{sec:res:inheritance_modeling}

\begin{figure}[!h]
\centering
\includegraphics[width=\linewidth]{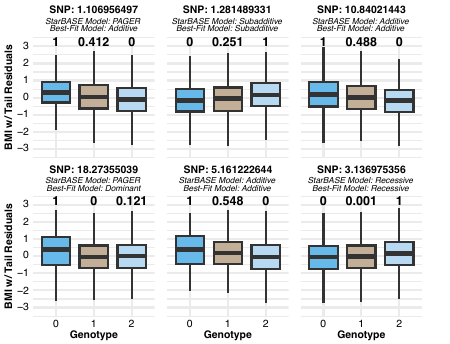}
\caption{Boxplots of BMI\_TAIL residuals across genotype classes for ground truth and putative QTLs. The most common StarBASE-GP encoder selection is listed above the best-fit inheritance model from the full dataset. Bolded values above each boxplot represent the normalized (between 0 and 1) observed mean phenotype per genotype class in the full dataset.}
\label{fig:inheritance-check}
\end{figure}

StarBASE-GP accurately encodes informative SNPs using inheritance models that either match the best-fit model or approximate it more closely when no strict model fits well by selecting PAGER (Figure \ref{fig:inheritance-check}). 
For two ground truth QTLs, $1.106956497$ and $18.27355039$, StarBASE-GP selects PAGER because the observed phenotype distributions fall between the assumptions of strict models. 
For example, at $18.27355039$, the normalized mean phenotype for genotype $2$ is $0.121$, whereas a dominant model would predict a value of $0$. 
PAGER captures this intermediate behavior more accurately, resulting in a higher marginal $r^2$ than the dominant encoding. 
Similarly, at $1.106956497$, the heterozygous genotype ($1$) deviates sufficiently from the additive expectation to warrant selection of PAGER over the additive model. 
These examples highlight StarBASE-GP’s ability to identify optimal encodings for each SNP, improving both interpretability and the accuracy of total genetic variance ($V_G$) by capturing additive ($V_A$) and dominance ($V_D$) components as needed. 

Across experiments, the distributions of optimal encoders for observed SNPs and Pareto front SNPs are diverse, with every encoding strategy represented (Supplementary File S1). 
Dominant, overdominant, and superadditive encodings consistently rank among the top selections across all experiments and SNP subsets, reflecting substantial $V_D$ in the dataset. 
Notably, the distribution of selected encoders shifts considerably between observed and Pareto SNPs in both StarBASE-GP and basic GP, with superadditive and additive models more common in Pareto SNPs--likely driven by strong signals at $1.281489331$ and $10.84021443$ (Figure \ref{fig:inheritance-check}).
PAGER was selected least frequently in all experiments and SNP subsets, underscoring its specialized but important role in modeling distributions that deviate from strict inheritance modes. 
Together, these findings further illustrate the prevalence of $V_D$ in real-world data and reinforce the need for flexible inheritance modeling approaches to accurately estimate $V_G$.
\section{Conclusions}
Compared to basic GP and the random control, StarBASE-GP consistently outperforms across multiple dimensions:
it achieves superior Pareto front characteristics (Figure\ref{fig:pareto-characteristics}), higher accuracy and precision in identifying ground truth targets (Tables\ref{tab:qtl_accuracy} and \ref{tab:qtl_positional_accuracy}), and stronger feature importance scores for both validated QTLs and novel candidates (Figure~\ref{fig:manhattan-SNPconsistency}). 
These gains stem from biologically inspired heuristics that reduce the search space and prioritize informative features.
Key contributors include LD-based pruning and the removal of uninformative SNPs (negative marginal $r^2$), which streamline feature selection and improve Pareto diversity. 
SNP DB further enhances pipeline construction by recommending high-value SNPs during smart mutation and crossover, balanced by random strategies to maintain exploration. 
In parallel, StarBASE-GP’s flexible encoding system enables accurate modeling of both additive ($V_A$) and dominance ($V_D$) effects (Figure \ref{fig:inheritance-check}), uncovering associations often missed by standard approaches.

We developed StarBASE-GP to efficiently analyze large-scale genomic data and detect SNPs with strong phenotypic associations through multiobjective evolutionary optimization. 
While GP provides a powerful foundation for exploring high-dimensional spaces, we show that success depends on the integration of biologically-informed heuristics. 
StarBASE-GP’s performance highlights the critical value of domain knowledge in GPA and positions it as a robust framework for uncovering meaningful genetic associations in complex traits and diseases.

\subsection{Limitations \& potential ways forward}
StarBASE-GP’s evolutionary search and requirement for replication result in substantially higher computational overhead than conventional GWAS.
While GWAS test each SNP individually under an additive encoding, they do not search for the optimal SNP subset that best explains the phenotype. 
In contrast, StarBASE-GP simultaneously optimizes SNP encodings and constructs informative subsets that jointly enhance phenotypic prediction.
Consequently, the computation time is justifiable, particularly when it leads to the discovery of novel, potentially valuable candidates. 
Nonetheless, reducing computation time remains an important future goal.

StarBASE-GP leverages multiprocessing to parallelize the pipeline evaluation process, enhancing its practical utility and enabling scalability.
Thus, a straightforward approach to reduce runtime is to increase computational resources.
However, efficiency gains are constrained by the available resources.
As an alternative direction, future work will explore the development of a GPU-accelerated version of StarBASE-GP. 
Notably, prior studies have demonstrated that GPUs can enhance the computational efficiency of evolutionary algorithms \cite{cheng2019gpu}.

StarBASE-GP’s computational demands scale with the number of variants, as the space of SNP combinations grows factorially. 
To manage this, the initial dataset can be partitioned and evaluated in parallel runs, retaining top-performing SNPs for cross-partition validation. 
Additionally, variants consistently flagged as uninformative can be excluded to further reduce dimensionality. 
Final StarBASE-GP runs can then focus on this refined SNP set for more targeted analysis.

\subsection{Future work}
\subsubsection{Extension to Binary Traits}
StarBASE-GP currently supports only continuous phenotypes. 
We will extend the framework to handle binary traits. 
This will involve implementing classification-based metrics for evaluating SNP and pipeline-level performances. 
Our existing biological heuristics can be easily adapted, making this a straightforward yet important step for broadening the tool’s applicability.

\subsubsection{Epistasis}
StarBASE-GP was developed with epistasis ($V_I$) in mind.  
However, extending the framework to model interactions ($n$ choose $k$, where $k > 1$) first requires validating its univariate capabilities ($k = 1$), which we demonstrate here.   
To manage the combinatorial complexity of modeling interactions, we will incorporate dimensionality reduction techniques such as multifactor dimensionality reduction (MDR)\cite{ritchie2001multifactor} and draw on principles from network science.
This extension will enable StarBASE-GP to explore complex genetic architectures involving multi-locus interactions, further enhancing its ability to describe ($V_G$) beyond single-locus effects.

\subsubsection{Leveraging external knowledge systems}
Generative AI, agentic AI, and knowledge graphs are emerging innovations that can enhance search capabilities and feature selection\cite{romano2024alzheimer, matsumoto2024kragen, matsumoto2025escargot}.  
We aim to develop StarBASE extensions that enable communication with and construction of external knowledge databases for iterative search refinement.
These technologies offer promising alternatives to brute-force search by enabling efficient traversal of large, complex feature spaces.

\subsubsection{GP configuration exploration}
The configuration of the GP system in StarBASE-GP is central to its effectiveness. 
Future work will explore how varying operator parameters influence outcomes. 
For example, tuning the ratio of smart to random sampling may improve efficiency, though over-reliance on smart strategies could lead to premature convergence to locally optimal solutions.

\section*{Data and code availability}

The data and source code for replication are available at \url{https://osf.io/b5dqm/}.
StarBASE-GP is an open source project available at \url{https://github.com/EpistasisLab/StarBASE-GP}.

\section*{Acknowledgments}
We thank the Department of Computational Biomedicine at Cedars-Sinai Medical Center for providing high-performance computing resources.
The work was supported by NIH grants R01 LM010098, U01 AG066833, and R01 LM014572 awarded to JHM.

 \bibliographystyle{IEEEtran}
 \bibliography{references}

% \newpage

 

% \end{IEEEbiography}



% \vfill

\end{document}